# Evaluating the Application of ChatGPT in Outpatient Triage Guidance: A Comparative Study

Dou Liu[1], Ying Han[2], Xiandi Wang[2], Xiaomei Tan[1], Di Liu[1,4], Guangwu Qian[3,4], Kang Li[1,4], Dan Pu[2], and Rong Yin[1,4(✉)]

[1] Department of Industrial Engineering, Pittsburgh Institute, Sichuan University, 610207 Chengdu, China
[2] Department of Medical Simulation Center, West China Hospital, Sichuan University, 610041 Chengdu, China
[3] Department of Computer Science, Pittsburgh Institute, Sichuan University, 610207 Chengdu, China
[4] West China Biomedical Big Data Center, West China Hospital, Sichuan University, 610041 Chengdu, China
`Corresponding author rong.yin@scupi.cn`

**Abstract.** The integration of Artificial Intelligence (AI) in healthcare presents a transformative potential for enhancing operational efficiency and health outcomes. Large Language Models (LLMs), such as ChatGPT, have shown their capabilities in supporting medical decision-making. Embedding LLMs in medical systems is becoming a promising trend in healthcare development. The potential of ChatGPT to address the triage problem in emergency departments has been examined, while few studies have explored its application in outpatient departments. With a focus on streamlining workflows and enhancing efficiency for outpatient triage, this study specifically aims to evaluate the consistency of responses provided by ChatGPT in outpatient guidance, including both within-version response analysis and between-version comparisons. For within-version, the results indicate that the internal response consistency for ChatGPT-4.0 is significantly higher than ChatGPT-3.5 ($p=0.03$) and both have a moderate consistency (71.2% for 4.0 and 59.6% for 3.5) in their top recommendation. However, the between-version consistency is relatively low (mean consistency score=1.43/3, median=1), indicating few recommendations match between the two versions. Also, only 50% top recommendations match perfectly in the comparisons. Interestingly, ChatGPT-3.5 responses are more likely to be complete than those from ChatGPT-4.0 ($p=0.02$), suggesting possible differences in information processing and response generation between the two versions. The findings offer insights into AI-assisted outpatient operations, while also facilitating the exploration of potentials and limitations of LLMs in healthcare utilization. Future research may focus on carefully optimizing LLMs and AI integration in healthcare systems based on ergonomic and human factors principles, precisely aligning with the specific needs of effective outpatient triage.

## 1 Introduction

In recent years, the rapid evolution of Large Language Models (LLMs) has significantly advanced ability in reading and generating human language texts. Among these, ChatGPT has emerged as a particularly notable model [1], attracting widespread attention upon its release. Attributed to its profound capabilities in handling text-based problems, ChatGPT has been extensively integrated across various domains, especially in healthcare [2–7]. Exploring the integration of Artificial Intelligence (AI) in healthcare settings and hospital systems stands as a critical frontier for groundbreaking innovations [8]. As a model underpinned by an extensive database, it can easily make a quick analysis of given symptoms [9] and generate highly accurate information in response to diverse medical queries [10]. Several teams have demonstrated the flexibility of ChatGPT 4.0, showcasing its capacity to not only successfully tackle questions from the United States Medical Licensing Exam but also achieve a high percentile in the Clinical Knowledge section of the Chinese Medical Licensing Examination, which is estimated to be in the top 20% [11, 12]. Furthermore, its utility extends to aiding in clinical decision-making [13–15], compiling medical documentation [16–18], and facilitating diagnostic processes [19–24].

Hospitals, as a critical sector in healthcare, require efficient information processing power and highly accurate and consistent responses regarding medical inquiries. Two key components, emergency department and outpatient services, both include a Pre-treatment step (triage in emergency and department guidance in outpatient), where their efficiency and accuracy significantly impact patient outcomes and the rational allocation of healthcare resources. Challenges such as prolonged wait time, complex diagnostic procedures, and fragmented communication channels necessitate a reevaluation of existing workflows. Consequently, applying ChatGPT, grounded in ergonomic and human factors engineering principles, to emergency triage and outpatient guidance in hospitals has the potential to swiftly provide judgments, address the challenges of medical resources shortage and high pressure. ChatGPT has already demonstrated excellent judgment in emergency triage decisions. A recent study observed a near-perfect concordance between the triage team's decisions and the established gold standard, as well as between ChatGPT-4's recommendations and that gold standard [25]. Another study highlighted ChatGPT-4's proficiency in making patient admission or discharge decisions in cases of metastatic prostate cancer in emergency departments, demonstrating a sensitivity of 95.7% and a specificity of 18.2% compared to the judgments of emergency physicians [26]. However, as another crucial function in hospitals, outpatient departments have received less research attention regarding the integration of LLMs.

In recent years, the internet has emerged as a critical health resource. Reports have indicated that up to 80% of users have explored health information online [27], and around 60% of patients reported investigating their symptoms online before or after visiting a doctor [28]. Furthermore, the potential for integrating AI into outpatient services is highlighted by public receptivity to digital assistance, as a previous study suggests most internet users are open to using health chatbots (with an acceptability rate of 67%) [29]. Nevertheless, research concerning the integration of AI into outpatient services are limited. Given this context, the deployment of ChatGPT in outpatient guidance is not only timely but could also substantially enhance operational efficiency across outpatient departments. By utilizing the tendency of patients to engage with digital platforms for health information, ChatGPT can serve as a critical bridge between online health searches and structured medical guidance, potentially optimizing the patient experience and improving outpatient services.

To understand ChatGPT's performance in Chinese outpatient guidance, this study aims to specifically assess the consistency of ChatGPT 3.5 and 4.0's judgment in Chinese outpatient settings through within-version analysis and between-version comparisons. The results are expected to examine the influence of the integration of LLMs in outpatient operations.

## 2 Methods

### 2.1 Study Design

This study employed a comparative design to assess the consistency of response generated by ChatGPT-3.5 and ChatGPT-4.0 in the context of Chinese outpatient guidance. We prepared a pre-tested prompt for ChatGPT (both 3.5 and 4.0 versions) to ensure they understood the required information. Fifty-two questions in Chinese encapsulating verbalized symptom descriptions, all of which were randomly collected from Xiaohongshu, a widely used Chinese social media, and Dingxiang Doctor, a Chinese online medical platform. To achieve the goal of testing the between-version and within-version consistency, each question was respectively inputted into both versions of ChatGPT three times. Considering the memory bias of LLMs, every question was treated as a new chat independently, instead of consecutive entries. The responses were carefully cataloged and recorded, noting the top three recommended outpatient departments ranked by probability (in total 100% for 3 recommended outpatient departments according to the provided symptoms), along with the probability of symptom recognition and being cured in each recommended outpatient department. This design allowed us to conduct a thorough within-version response analysis and a between-version comparison, laying the foundation for a comprehensive evaluation of the consistency and reliability of AI-assisted outpatient guidance.

### 2.2 Data Collection

A total of 52 questions generated 156 responses, which were each assessed for consistency between the responses for a same question. The ineffective responses (such as 0 or Nan) were excluded to ensure the validity of data. Additionally, when calculating consistency, the order of recommended departments was excluded from consideration. The data collection was conducted between January 2024 to March 2024.

The within-version consistency of responses to individual questions was closely monitored. Each set of three responses, corresponding to the same question and the same GPT version, was evaluated based on the two-tier grading system designed as follows: (1) Completely inconsistency, where any two of the three responses are not exactly the same. (2) Partially consistent or completely consistent, where there are at least two responses containing identical recommended departments.

In addition to the consistency within the three responses, significant attention was dedicated to the consistency of the top recommended department for each question, which is also a key aspect of internal consistency. We used another two-tier grading system similar to the aforementioned one: (1) Partially consistent/inconsistent, where no more than two identical department was mentioned among the three responses. (2) Completely consistent, where all three responses conveyed the same recommendation.

Additionally, the between-version consistency of the responses from GPT3.5 and GPT4.0 was taken into consideration. A horizontal comparison was conducted using a three-point grading system: (1) 0-point—incoherence, where there are no matching recommended outpatient departments between the two versions' responses. (2) 1-point – partially consistent, where only one department is common between the two responses. (3) 2-points – mostly consistent, where there is only one inconsistent department. (4) 3-points – completely consistent, all three recommendations are identical.

### 2.3 Data Analysis

The data analysis was conducted in Python, focusing on examining the differences in within- and between-version response consistency between ChatGPT-3.5 and ChatGPT-4.0. To determine if there are statistically significant differences in responses across two versions, we employed the Chi-square test. A significance level of $\alpha=0.05$ was set. This analysis aimed to quantify the extent of consistency between the two versions in both within- and between- version comparisons.

## 3 Results

### 3.1 Completeness and recommendation probability

The statistical descriptives are presented in **Table 1**. Of all the recorded responses, only 6 (1.9%) out of

312 were categorized as invalid or ineffective, lacking critical elements such as recommendations. Seventy-one (22.8%) responses were incomplete, missing elements such as probability, urgency degree, or recommended departments (26 responses from ChatGPT-3.5 and 45 responses from ChatGPT-4.0). The completeness rate of response from ChatGPT-3.5 was statistically significantly higher than that of ChatGPT-4.0 ($p = 0.02$).

The proportion that the top-recommended department has a Recommendation Rate (RR) of higher than 50% is 58.4% (179 out of 306 responses), among which 20 (20 out of 179 responses, 11.2%) responses had a RR higher than 80% for the top recommendation. Additionally, the probability that the top-recommended outpatient department can successfully treat the symptom is 61.1% (187 out of 306 responses), among which 11 (11 out of 187 responses, 5.9%) responses had a probability higher than 80% that the symptom would be successfully treated.

**Table 1.** Descriptive Statistics

| Description | Quantity (n, %) |
| --- | --- |
| Total number of responses | 312 (100%) |
| Valid responses | 306 (98.1%) |
| Invalid responses | 6 (1.9%) |
| Complete output | 241 (77.2%) |
|     Complete output from ChatGPT-3.5 | 130 (41.7%) |
|     Complete output from ChatGPT-4.0 | 111 (35.6%) |
| Incomplete output | 71 (22.8%) |
|     Incomplete output from ChatGPT-3.5 | 26 (8.3%) |
|     Incomplete output from ChatGPT-4.0 | 45 (14.4%) |
| The top recommended department with RR>80% | 20 (6.5%) |
| The top recommended department with 80%≥ RR≥50% | 179 (58.4%) |
| The top recommended department with RR < 50% | 127 (41.5%) |
| Top recommended department can treat the symptom with probability P> 80% | 11 (3.6%) |
| Top recommended department can treat the symptom with probability 80%≥P≥ 50% | 187 (61.1%) |
| Top recommended department can treat the symptom with probability P< 50% | 119 (38.9%) |

### 3.2 Within-version consistency

**Table 2** represents the results of internal consistency of responses to individual questions. The total number of questions analyzed for both ChatGPT-3.5 and 4.0 was 52. For ChatGPT-3.5, there were 20 (38.5%) completely inconsistent sets of responses among the 52 questions, while ChatGPT-4.0 exhibited only 9 (17.3%) completely inconsistent sets. ChatGPT-4.0 demonstrated higher internal consistency in the response sets for each question ($p = 0.03$).

Table 2. Within-version consistency of responses

| LLMs | Total number of questions | Completely inconsistent | Partially /Completely consistent |
|---|---|---|---|
| ChatGPT-3.5 | 52 | 20 | 32 |
| ChatGPT-4.0 | 52 | 9 | 43 |

**Table 3** summarizes the consistency of the top recommendation in each set of responses. We observed that 31 (59.6%) sets of responses had completely consistent recommendations in ChatGPT-3.5 and 37 (71.2%) sets in ChatGPT-4.0. No statistically significant differences were found for the within-version consistency for the top recommendation within the two ChatGPT versions (p>0.05). In this case, both versions have the ability to deliver responses with moderate consistency.

Table 3. Within-version consistency of top recommendation

| LLMs | Total number of questions | Completely consistent | Partially consistent/inconsistent |
|---|---|---|---|
| ChatGPT-3.5 | 52 | 31 | 21 |
| ChatGPT-4.0 | 52 | 37 | 15 |

### 3.3   Between-version consistency

The between-version consistency of the responses between ChatGPT versions 3.5 and 4.0 is presented in **Fig. 1**. A descriptive analysis was conducted: after removing the invalid data, 150 paired responses remained. The mean score is 1.43, with 13 pairs (8.7%) scored 0 point, 69 pairs (46%) scored 1 point, 59 pairs (39.3%) scored 2 points, and 9 pairs (6%) scored 3 points. The median score of the entire dataset was 1, indicating that there is a moderate level of match between the responses from versions 3.5 and 4.0. Regarding the top recommendation consistency, 50% of the responses (75 out of 150) matched each other, with a mean score of 0.5 out of 1 and a median of 0.5.

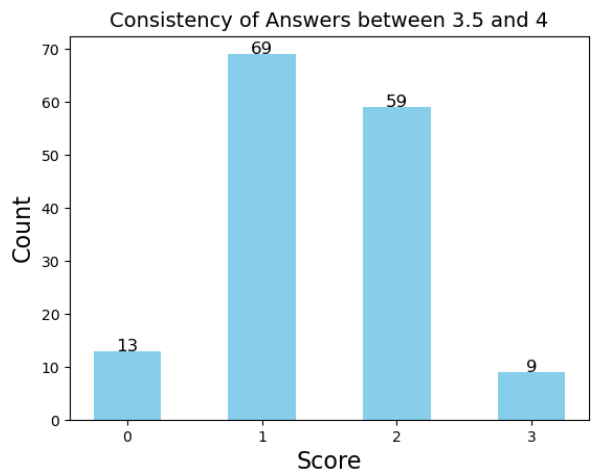

Fig. 1. Consistency between responses of ChatGPT-3.5 and ChatGPT-4.0.

# 4    Discussion

In outpatient settings, efficient and accurate triage is critical to patient satisfaction. This study explored the potential of using ChatGPT as a simulated AI chatbot in the outpatient triage process, which effectively mirrored real-world outpatient guidance, contrasting with prior research focused mainly on emergency department triage accuracy [30]. Our findings suggest that ChatGPT could aid in outpatient guidance, with ChatGPT-4.0 showing superior internal consistency, potentially offering stable responses to patient-reported symptoms. The integration streamlines the workflow in outpatient services, reducing wait times, and facilitating more consistent and timely medical recommendations. However, both ChatGPT versions only moderately align in generating consistent top recommendations, highlighting the risk of inaccurate advice in practical applications. Additionally, the between-version response comparison revealed low consistency, suggesting significant variability in accuracy between ChatGPT-3.5 and ChatGPT-4.0.

In our study, changing prompts led to various responses, highlighting the significance of prompt engineering in LLMs-based chatbots. A recent study has shown the importance of precise prompts and highlighted prompt skill as one emerging skill for medical professionals [31]. Effective LLM utilization involves inputting appropriate prompts to obtain accurate responses. Thus, tailored prompts could direct AI responses more effectively in healthcare, regarding improved accuracy and efficacy. The potential of prompt engineering needs to be further explored in future studies.

Initially, we input questions consecutively, leading to AI's memory bias, as responses started referencing 'updated symptoms,' merging prior and current queries. To counteract this, we adjusted our method to independently re-enter each question three times in new chats. The memory property of AI should be considered in such evaluations. Meanwhile, differences in response speed between ChatGPT-3.5 and 4.0 were observed, with speed ratios reaching 6:1. Despite 4.0's higher consistency, its slower response may affect operational efficiency, especially under high time pressure scenarios.

This research has several limitations. First, the questions were posed in Chinese, not English, the original language of ChatGPT. Although ChatGPT now supports communications in multiple languages including Chinese, Spanish, and some others, there remains a concern that the generated content may not perfectly convey the intended meaning when using languages other than English. In our study, we observed that ChatGPT provided different Chinese translations for the same outpatient department named in English. However, this discrepancy does not affect language fluency, which aligns with a previous study focusing on the performance of ChatGPT in a Chinese context for a medical task, where verbal fluency exceeded 95% [32]. Future studies may explore how the language used to interact with ChatGPT influences information processing and response generation.

The integrity of our consistency analysis depended on question quality. Poorly described symptoms resulted in inconsistent outcomes, emphasizing the importance of detailed, precise symptom information to improve AI's diagnostic accuracy. Future studies may consider description patterns to refine question description methodologies, ensuring that workflow optimization does not come at the expense of comprehensive and personalized patient care.

# 5    Conclusion

This study examined the feasibility of applying ChatGPT in outpatient triage, demonstrating its potential to augment medical decision-making. While indicating the need for further AI refinement, our findings particularly spotlight the varied performance between different versions, suggesting a tailored approach to its application in healthcare. Future research may focus on carefully optimizing LLMs and AI integration in healthcare systems, precisely aligning with the specific needs of effective outpatient triage.